\documentclass[submission,copyright,creativecommons]{eptcs}
\providecommand{\event}{DCM} 
\usepackage[utf8]{inputenc}
\usepackage[T1]{fontenc}
\usepackage{textcomp}
\usepackage[english]{babel}
\usepackage{amsmath, amssymb}
\usepackage{tikz}
\usepackage{xfp, siunitx}
\usepackage{pgfplots}

\usepackage{gb4e}
\noautomath 

\usepackage{import}
\usepackage{xifthen}
\usepackage{pdfpages}
\usepackage{transparent}

\newtheorem{definition}{Definition}

\usetikzlibrary{arrows,positioning,matrix,backgrounds,calc,fit,decorations.pathreplacing,intersections,through}
\usepgfplotslibrary{statistics, fillbetween}
\pgfplotsset{compat=newest}
\pgfplotsset{height=4cm, width=7cm}
\pgfplotsset{
  layers/axis lines on top/.define layer set={
    axis background,
    axis grid,
    axis ticks,
    axis tick labels,
    pre main,
    main,
    axis lines,
    axis descriptions,
    axis foreground,
  }{/pgfplots/layers/standard},
}

\newcommand{\histogram}[3][]{
\begin{tikzpicture}
\begin{axis}[
    ymin=0,
    xmin=0, xmax=1,
    xtick={0,1/6,2/6,3/6,4/6,5/6,1},
    xticklabels={$0$,  $\frac{1}{6}$,  $\frac{2}{6}$,  $\frac{3}{6}$,  $\frac{4}{6}$,  $\frac{5}{6}$,  $1$},
    xlabel={Signalling fraction},
    minor y tick num = 4,
    grid,
    grid style=dashed,
    legend pos={outer north east},
    legend cell align={left},
    ]
\addplot+[ybar interval,mark=no,color=blue,fill=blue,fill opacity=0.3,ybar legend] plot coordinates {
#2 (0.16666667, 0)
};
\addplot+[ybar interval,mark=no,color=red,fill=red,fill opacity=0.3,ybar legend] plot coordinates {
#3 (1, 0)
};
\ifthenelse{\isempty{#1}}
{\legend{contextual, inconclusive}}
{}
\end{axis}
\end{tikzpicture}
}

\newcommand{\cbdhistogram}[3][]{
\begin{tikzpicture}
\begin{axis}[
    ymin=0,
    xmin=0, xmax=6,
    xtick={0,1,2,3,4,5,6},
    xticklabels={$0$,  $1$,  $2$,  $3$,  $4$,  $5$,  $6$},
    xlabel={Direct influence},
    minor y tick num = 4,
    grid,
    grid style=dashed,
    legend pos={outer north east},
    legend cell align={left},
    ]
\addplot+[ybar interval,mark=no,color=purple,fill=purple,fill opacity=0.3,ybar legend] plot coordinates {
#2 (2, 0)
};
\addplot+[ybar interval,mark=no,color=brown,fill=brown,fill opacity=0.3,ybar legend] plot coordinates {
#3 (6, 0)
};
\ifthenelse{\isempty{#1}}
{\legend{contextual, non-contextual}}
{}
\end{axis}
\end{tikzpicture}
}

\newcommand{\cnt}[1]{{\textsf{CNT}\text{$_#1$}}}
\newcommand{\cf}{\textsf{CF}}
\newcommand{\csf}{\textsf{SF}}

\begin{document}
\title{Developments in Sheaf-Theoretic Models of Natural Language Ambiguities}
\def\titlerunning{Sheaf Theory and Natural Language}

\author{Kin Ian Lo \qquad Mehrnoosh Sadrzadeh
    \institute{University College London \\London, UK}
    \email{\{kin.lo.20,m.sadrzadeh\}@ucl.ac.uk}
    \and
    Shane Mansfield
    \institute{Quandela \\ Paris, France}
    \email{shane.mansfield@quandela.com}
}

\def\authorrunning{K. I. Lo, M. Sadrzadeh \& S. Mansfield}

\maketitle 

\begin{abstract}
    Sheaves are mathematical objects consisting of a base that constitutes a topological space and the data associated with each open set thereof, e.g.\ continuous functions defined on the open sets. Sheaves have originally been used in algebraic topology and logic. Recently, they have also modelled events such as physical experiments and natural language disambiguation processes. We extend the latter models from lexical ambiguities to discourse ambiguities arising from anaphora. 
    To begin, we calculated a new measure of contextuality for a dataset of basic anaphoric discourses, resulting in a higher proportion of contextual models--82.9\%--compared to previous work which only yielded 3.17\% contextual models.
    Then, we show how an extension of the natural language processing challenge, known as the Winograd Schema, which involves anaphoric ambiguities can be modelled on the Bell-CHSH scenario with a contextual fraction of $0.096$.
\end{abstract}

\section{Background}

We introduce the basic elements of presheaf and sheaf theory, review sheaf theoretic models of quantum scenarios and of lexical ambiguities of natural language.

\subsection{Sheaf Theory}

The \emph{presheaf} of events is a contravariant functor ${\cal E}$ from subsets of a set $X$ to the category $Set$:

\[
    {\cal E} \colon {\cal P}(X)^{op} \to Set
\]
Given a set $O$ of outcomes, the functor acts as follows on objects, i.e. subsets $U$ of $X$:
\[
    {\cal E} \colon U \mapsto O^U
\]
Given $U, U' \subseteq X$ and for $res^{U'}_{U}$ the restriction map, ${\cal E}$ acts as follows on morphisms:

\begin{eqnarray*}
    &&U \subseteq U' \implies res^{U'}_U \colon {\cal E}(U') \to {\cal E}(U) :: s \mapsto s|U\\
    && U \subseteq U' \subseteq U'' \implies res^{U'}_U \circ res^{U'}_U = res^{U''}_U
\end{eqnarray*}

\noindent
That is, it assigns to each subset $U \subseteq X$ a map $s \colon U \to O$. This map describes the event in which an element $m \in U$ is performed with the outcome $s(m)$. It describes a section of the data that is being modelled and is thus referred to as a \emph{section}.
A presheaf is a \emph{sheaf} if every compatible family of sections can be glued together to form a global section.
This means that for some locally compatible sections $\{s_i \in {\cal E}(U_i)\}_{i \in I}$, there is a unique global section $s \in {\cal E}(U)$ such that $s|U_i = s_i$ for all $i \in I$. In other words, \emph{local} data must agree on overlaps. If this is the case, we can \emph{glue} it together to create a (unique) \emph{global} section $s \in {\cal E}(U)$ such that $s|U_i = s_i$, for all $i \in I$.

We can assign to each set of measurements $U$, the set of probability distributions over the sections of $U$ by composing the sheaf functor ${\cal E}$ with a distribution functor ${\cal D}_R$. This provides us with the functor ${\cal D}_R({\cal E}(U))$. The distribution functor is over a semiring $R$. It assigns to a set $X$ of measurements, a set of functions $d \colon X \to R$ with finite support, such that $\sum_{x \in X} d(x) = 1$. The resulting composed functor has the type ${\cal D}_R {\cal E} \colon {\cal P}(X)^{op} \to Set$ and is a presheaf, as for $U \subseteq U'$, we can define a restriction map as follows:

\[
    {\cal D}_R {\cal E}(U') \to {\cal D}_R {\cal E}(U) :: d \mapsto d|U
\]
This map sends a section $s \in {\cal E}(U)$ to its marginal $d|U(s)$, that is the sum of all probabilities in the larger sections $d(s')$ for $s' \in {\cal E}(U')$ whenever those larger sections $s'$ restricted to $s$, i.e.\ $s' \mid U = s$.

\subsection{Sheaf Models of Quantum Scenarios}

\begin{figure}[t!]
    \centering
    (a)
    \fbox{
        \begin{tikzpicture}[x=45pt,y=45pt,thick,label distance=-0.25em,baseline=(O.base), scale=0.7]
    \def\basenameone{$a_1$}
    \def\basenametwo{$b_1$}
    \def\basenamethree{$a_2$}
    \def\basenamefour{$b_2$}

    \def\outcomeone{$0$}
    \def\outcometwo{$1$}

    \coordinate (O) at (0,0);
    \def\sectionoffset{1.5}
    \def\sectionsep{0.5}

    \foreach \i in {0,1} {
        \coordinate (t\i) at (0, \sectionoffset + \sectionsep*\i);
    }

    \foreach \i/\p in {0/0,1/6,2/12,3/18} {
        \coordinate [inner sep=0em] (v\i) at ($ (
            {-cos(\p*pi/12 r)*0.8},
            {-sin(\p*pi/12 r)*0.8}
            ) $);
        \foreach \j in {0,1} {
            \coordinate [inner sep=0em] (v\i-\j) at ($ (v\i) + (t\j) $);
        }
    }

    \foreach \i/\j in {0/1,1/2,2/3,3/0} {\draw (v\i) -- (v\j);}


    \node [inner sep=0.1em,label={[label distance=-0.25em]left:{\basenameone}}] at (v0) {$\bullet$};
    \node [inner sep=0.1em,label={[label distance=-0.625em]330:{\basenametwo}}] at (v1) {$\bullet$};
    \node [inner sep=0.1em,label={[label distance=-0.25em]right:{\basenamethree}}] at (v2) {$\bullet$};
    \node [inner sep=0.1em,label={[label distance=-0.5em]175:{\basenamefour}}] at (v3) {$\bullet$};

\end{tikzpicture}
    }
    \qquad
    (b)
    \begin{tabular}{r|ccccc}
                     & $(0, 0)$ & $(0, 1)$ & $(1, 0)$ & $(1, 1)$ \\ \hline
        $(a_1, b_1)$ & $1 / 2$  & $0$      & $0$      & $1 / 2$  \\
        $(a_1, b_2)$ & $3 / 8$  & $1 / 8$  & $1 / 8$  & $3 / 8$  \\
        $(a_2, b_1)$ & $3 / 8$  & $1 / 8$  & $1 / 8$  & $3 / 8$  \\
        $(a_2, b_2)$ & $1 / 8$  & $3 / 8$  & $3 / 8$  & $1 / 8$  \\
    \end{tabular}
    \label{fig:basechsh}
    \caption{The geometric representation of the Bell-CHSH scenario and an empirical model for it.}
\end{figure}
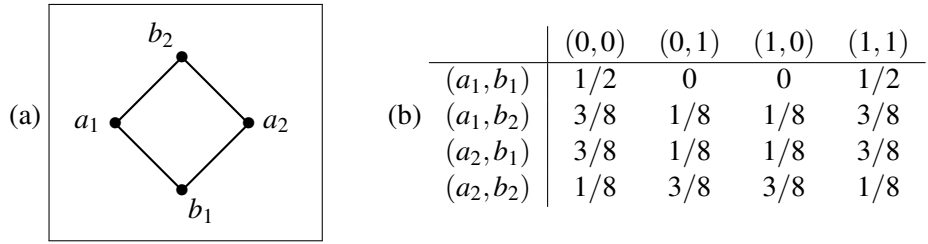
An example of a sheaf of events with distributions is the entanglement protocols of quantum mechanics. Here, we have measurement scenarios, which are modelled by tuples of the form $\langle X, \mathcal{M}, {O} \rangle$, where $\cal X$ is the set of observables of the scenario, $\cal M$ is a cover of $X$, and $O$ is the set of measurement outcomes. A subset of simultaneously measurable observables of ${X}$ is called a measurement context.
%
%
Thus, one can define a (local) joint probability distribution over the observables of a context.
Such a joint probability distribution can either be estimated by performing the measurements in an experiment, or by calculating according to the theory of the system of interest. A collection of all joint probability distributions is called an \emph{empirical model}.

Sheaf theory can be used to model and reason about a fundamental phenomenon of quantum mechanics known as \emph{contextuality}. This phenomenon was observed as early as 1935 by Einstein, Podolsky and Rosen (EPR)~\cite{EPR}. It occurs when the quantum mechanical description of a scenario is incomplete.
A typical scenario that exhibits contextuality has two space-like separated parties allowed to make measurements on an entangled system. In a sheaf theoretic setting, contextuality becomes the fact that there does not exist a global joint distribution over all the observables of the scenario that marginalises to every local joint distribution in the empirical model. This means failure of finding a classical explanation for the empirical model of the scenario. In conclusion, a scenario can host any contextual model if and only if its distribution presheaf is not a sheaf.

As an example, consider the Bell-CHSH scenario~\cite{Clauser1969}. It is specified by
$X = \{a_1, a_2, b_1, b_2\}$; $\mathcal{M} = \big\{ \{a_1, b_1\}, \{a_1, b_2\}, \{a_2, b_1\}, \{a_2, b_2\} \big\}$; $O = \{0, 1\}$. $\mathcal{M}$ has a geometric realisation as the boundary of a square (a cycle of order 4), where each vertex is an observable, and each edge is a context (see Figure~\ref{fig:basechsh}). The Bell state $|\Psi \rangle = \big(|00\rangle + |11\rangle\big)/\sqrt{2}$ produces the empirical model shown in Figure~\ref{fig:basechsh}, which is contextual and moreover violates the Bell-CHSH inequality maximally~\cite{Bell1964,Clauser1969}.


An empirical model is \emph{signalling} if the marginalised distribution of a set of observables differs from one context to another.
Intuitively speaking, non-signalling means that the observed probabilities are context-invariant and thus the choice of context cannot be used to transmit information over parties.
A degree of contextuality can be defined for a model using its \emph{contextual fraction} $\text{\cf}(e)$, introduced in~\cite{Abramsky2017}. Given an empirical model $e$, the contextual fraction $\text{\cf}(e)$ is the minimum $\lambda$ such that $e$ admits a convex decomposition:
\begin{equation}
    \label{eq:convexcf}
    e = (1-\lambda) e^{NC} + \lambda e^{C},
\end{equation}
where $e^{NC}$ is a non-contextual empirical model and $e^{C}$ is an empirical model that may be contextual. Suppose a given model $e$ is non-contextual, then $\lambda$ can be set to zero by choosing $e^{NC} = e$. Otherwise, $\lambda$ must be greater than zero to make the decomposition valid. In a nutshell, for all contextual models, we have:
\begin{equation}
    \label{eq:cfnosig}
    \text{\cf}(e) > 0
\end{equation}
The {\cf} of a model has a nice interpretation as the maximum amount of \emph{normalised violation} of all possible general Bell's inequalities~\cite{Abramsky2017}. For signalling models, the above decomposition would never hold as $e^{NC}$ and $e^{C}$ are non-signalling by definition.
We could instead allow $e^{C}$ to be signalling, but doing so would lead to the erroneous conclusion that all signalling models are contextual, assuming we still interpret {\cf} as a measure of contextuality for signalling models.

Emeriau et al.~\cite{Emeriau2022} proposed a criterion for contextuality in the presence of signalling, making use of a measure called the \emph{signalling fraction} $\text{\csf}(e)$, defined as the minimum $\lambda$ such that $e$ admits a convex decomposition:
\begin{equation}
    \label{eq:convexsf}
    e = (1-\lambda) e^{NS} + \lambda e^{S},
\end{equation}
where $e^{NS}$ is a non-signalling empirical model and $e^{S}$ is an empirical model that may be signalling. The signalling fraction of a model is zero if and only if the model is non-signalling. 
The contextuality criterion of Emeriau requires the following inequality to hold:
\begin{equation}
    \label{eq:sfcriterion}
    \text{\cf}(e) > 2 |\mathcal{M}| \csf(e).
\end{equation}
The intuition is that the degree of signalling (\csf) can be regarded as the magnitude of perturbation on the empirical model. The perturbation in turn affects the contextual fraction (\cf) of the model. 
Emeriau et al. proved a continuity result that bounds the change to the contextual fraction given the magnitude of a perturbation of the empirical model. 
The factor of $|\mathcal{M}|$ comes from the fact that every context in the scenario could be perturbed independently, and thus their effects could accumulate. The factor of $2$ comes from the use of \emph{total variation} as a distance measure between empirical models. The total variation distance between two probability distributions is defined as half the sum of the absolute differences between their probabilities.

\subsection{Sheaf Models of Lexical Ambiguities}

Lexical ambiguity, where a word has multiple meanings, is one of the most common types of ambiguity in natural language. One way to formalise lexical ambiguity in natural language is to consider an ambiguous word as an observable. The possible interpretations of the ambiguous word are treated as the possible outcomes of the observable. As an example, consider the ambiguous word \emph{pitcher}, which means \emph{a certain type of jug} or \emph{or a type of baseball player}. These interpretations are modelled as outcomes. Reading the word \emph{pitcher} in a piece of text thus becomes the event of performing a measurement. The word itself becomes an observable. Probabilities are assigned to outcomes by asking a group of readers to read a word and rate the likelihood of its interpretations over one another. This way of treating ambiguous words is inspired by the Bell-CHSH scenario and was first considered in~\cite{Wang2021a,Wang2021b}. The authors considered subject-verb and verb-object phrases where each word in the phrase had at least two possible interpretations. The measurement contexts were constructed by selecting different pairs of nouns and verbs in a way similar to how Alice and Bob select their measurements in the Bell-CHSH scenario. The phrases were uploaded into the crowd-sourcing Amazon Mechanical Turk and probability judgments were collected from human subjects.
The resulting empirical models were all signalling and the authors had to analyze the data in the framework of Contextuality-by-Default~\cite{Dzhafarov2013,Dzhafarov2016a,Dzhafarov2015a}. This setting is able to compute a degree of contextuality in the presence of signalling. According to this measure,  $4.5\%$ of the phrases (13 out of 290) in~\cite{Wang2021b} were contextual.

\section{Methodology}
In this section, we introduce the basic anaphoric ambiguity schema and the Winograd Schema. We then describe how we modelled them using the sheaf framework and the Contextuality-by-Default framework.

\subsection{Sheaf Models of Basic Anaphoric Ambiguities}
\label{subsec:prprism}
Another type of ambiguity in natural language is \textit{anaphoric ambiguity}. Here, a pronoun can refer to different expressions that come before it in a piece of text.
For instance, the pronoun \textit{it} (aka \emph{anaphora}) can refer to the \textit{dog} or the \textit{cat} (aka \emph{anaphors}) in the text \textit{The dog chased the cat. It barked.}. A piece of text that contains an anaphora together with its possible anaphors is often called a \emph{discourse}.

In a previous work~\cite{Lo2022}, we discovered that a simpler scenario than Bell-CHSH can be used to model basic anaphoric ambiguities.
This scenario has three (rather than four) observables ${ \cal X} = \{x_1, x_2, x_3\}$, which give rise to measurement contexts ${\cal M} = \{\{x_1, x_2\}, \{x_2, x_3\}, \{x_3, x_1\}\}$.
We proved that this scenario is the only strongly contextual scenario up to relabelling and called the models of the scenario \emph{PR prism}, as an analogy to the PR boxes.
We then built schemas demonstrating anaphoric ambiguities, where outcomes $O_1$ and $O_2$ were the choice of referents are noun phrases and the observables $X_1, X_2, X_3$ are any of their \emph{basic modifiers}, i.e. adjective, verb, and prepositional phrases. We refer to the ambiguities resulting from these constructions as \emph{basic anaphoric ambiguities}. 

We then built schemas in English demonstrating the basic anaphoric ambiguities. 
The schemas were instantiated using the large pre-trained transformer-based language model BERT~\cite{Devlin2019}. The masked word prediction capability of BERT  provided us with instances of the modifiers and their probability distributions.

\begin{figure}[t]
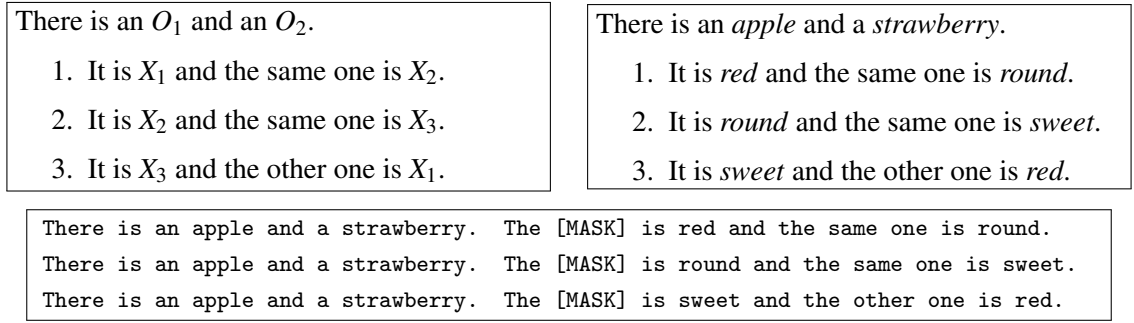

    \begin{center}
    \fbox{\begin{minipage}{7cm}{
     There is an $O_1$ and an $O_2$. 
     \begin{enumerate}
        \item It is $ X_1$ and the same one is $ X_2$. 
        \item It is $ X_2$ and the same one is $ X_3$. 
        \item It is $ X_3$ and the other one is $ X_1$.
     \end{enumerate}}\end{minipage}}
     \quad
     \fbox{\begin{minipage}{7cm}{
         There is an \emph{apple} and a \emph{strawberry}. 
     \begin{enumerate}
         \item It is \emph{red} and the same one is \emph{round}. 
         \item It is \emph{round} and the same one is \emph{sweet}. 
         \item It is \emph{sweet} and the other one is \emph{red}.
     \end{enumerate}}\end{minipage}}
 
     \vspace*{0.2cm}

     \fbox{
     \begin{minipage}{14cm}{
        {\footnotesize  \texttt{There is an apple and a strawberry. The [MASK] is red and the same one is round.}}\\
        {\footnotesize  \texttt{There is an apple and a strawberry. The [MASK] is round and the same one is sweet.}}\\
        {\footnotesize  \texttt{There is an apple and a strawberry. The [MASK] is sweet and the other one is red.}}
        }\end{minipage}
    }

     \end{center}
     \caption{The left box shows the general PR prism schema of basic anaphoric ambiguities. The two outcomes $O_1$ and $O_2$ are two nouns which are the referents of the anaphor. The three observables $X_1$, $X_2$, and $X_3$ are the modifiers of the nouns. The right box shows an instance of the schema with adjective modifiers.}
     \label{fig:schemadj}
     \end{figure}
    
In order to build discourses with basic anaphoric ambiguities, we used the template shown in Figure \ref{fig:schemadj} which involves two nouns ($O_1$ and $O_2$), and three modifiers ($X_1$, $X_2$, and $X_3$) which can be instantiated by any of the following types of modifier:
\begin{enumerate}
    \item adjectives, e.g.\ \emph{red}, \emph{round}, \emph{sweet}; (an example of which is shown in Figure \ref{fig:schemadj})
    \item participial adjectives (verb-derived adjectives), e.g.\ \emph{running}, \emph{broken}, \emph{frozen};
    \item prepositional phrases, e.g.\ \emph{in the box}, \emph{on the table}, \emph{under the bed}.
\end{enumerate}

In~\cite{Lo2022}, we constructed 11,052 examples of the schema with adjective modifiers. 
As the examples were in general signalling, we used the criterion of~\cite{Emeriau2022}, which makes use of the signalling fraction to determine if the examples were contextual. We found that only 350 examples (3.17\%) were contextual. 
Since the criterion of~\cite{Emeriau2022} is a lower bound, the actual percentage of contextual examples could be higher. 

\subsection{The Contextuality-by-Default (CbD) framework}
In the Contextuality-by-Default framework, one has two important notions: \emph{contents} $q_i$, which are questions about the system; and \emph{contexts} $c^j$, which represent the conditions under which the questions are asked. A question $q_i$ asked in a context $c^j$ gives rise to a random variable $R^j_i$ taking values in $\{\pm 1\}$, and representing possible answers and their probabilities. All random variables in a given context are jointly distributed.

The simplest types of CbD systems are $n$-cyclic. Here, each context has exactly 2 contents, and every content is exactly in 2 contexts. It has been proven in~\cite{Kujala2019} that these systems are contextual if and only if:
\begin{equation}\label{eq:BellInequality}
    \cnt{1} = \cnt{2} = s_{odd} \left(\left\{\left<R^{j}_{i_j}R^{j}_{i'_j}\right>\right\}_{j=1,\ldots,n}\right) - \Delta - n + 2 > 0
\end{equation}
where $j_i\neq j'_i$ for all $i$ and $R^{j}_{i_j}, R^{j}_{i'_j}$ are well-defined for all $j$. Quantities $s_{odd}: \mathbb{R}^n \to \mathbb{R}$ and $\Delta$ are defined as follows:
\begin{equation}
    s_{odd}\left(\underline{x}\right) = \max_{\substack{\underline{\sigma}\in \{\pm1\}^k; \\ \mathfrak{p}(\underline{\sigma}=-1)}}\underline{\sigma}\cdot \underline{x}\ ; \qquad
    \Delta = \sum_{i=1}^n \left|\left<R^{j_i}_{i}\right> - \left<R^{j'_i}_{i}\right>\right|
\end{equation}
where $\mathfrak{p}(\underline{\sigma}) = \prod_{i=1}^n \sigma_i$ ($\mathfrak{p}$ is the parity function of $\underline{\sigma}$).
The quantity $\Delta$ is called \emph{Direct Influence} and it measures the degree of signalling in the system (in a non-signalling system $\Delta=0$). In the Bell-CHSH scenario, $n=4$ and $S_{odd}$ becomes the maximum violation of the inequality.

Following Wang et al.~\cite{Wang2021a,Wang2021b}, instead of using the signalling fraction of sheaf theoretic models for our basic anaphoric ambiguities, we use the CbD framework.
We modelled our basic anaphoric ambiguity dataset, described in Section \ref{subsec:CbDbasicAnaphora}, by taking $q_i$ to be the anaphors, i.e.\ the adjective, verb, and prepositional phrase modifiers.
We took $c^j$ to be the different sentences in which the anaphors appear.

\subsection{Sheaf Theoretic Models of the Original Winograd Schema}
The Winograd Schema Challenge (WSC) consists of 285 descriptions with anaphoric ambiguities in them, followed by questions about the anaphors. The questions cannot be answered by syntactic reasoning only and extra information about the semantic context of the descriptions are needed. WSC was proposed as a test for human intelligence and an alternative to Turing's test in 2012~\cite{Levesque2012}.
An example description is \textit{The trophy doesn't fit into the suitcase because it's too [\textit{small} / \textit{large}].} The questions following this description are \textit{What is too [\textit{small} / \textit{large}]?} The correct answer for \textit{small} is \textit{suitcase}, and for \textit{large} is \textit{trophy} .
In 2014, a cash prize was put forwards for an AI that could solve the WSC with an accuracy close to humans. The prize was withdrawn in 2018 after failures of machine learning algorithms of the time. In 2019, however, the deep learning transformer algorithm RoBERTa~\cite{Liu2019} fine-tuned on large amounts of data, got close to solving it. Nevertheless, WSC still poses a challenge for AI's that do not have access to the large resources of these algorithms~\cite{Kocijan2020a}. In this subsection, we describe and model the challenge. A valid Winograd schema must satisfy the following requirements:
\begin{enumerate}
    \item It consists of a pair of sentences. The first sentence contains a \textit{special} word. The second sentence is obtained by replacing the \textit{special} word with an \textit{alternate} word. In the \textit{trophy-suitcase} example, the \textit{special} word is \textit{small} and the \textit{alternate} word is \textit{large}.

    \item There should be two noun phrases in the sentences. In the \textit{trophy-suitcase} example, the two noun phrases are \textit{the trophy} and \textit{the suitcase}.

    \item A pronoun must appear in the sentences.
          The pronoun must agree with the two noun phrases in terms of number and gender.
          In the \textit{trophy-suitcase} example, the pronoun is \textit{it} which agrees with both \textit{the trophy} and \textit{the suitcase} in terms of number and gender.

    \item The referent of the pronoun should be easily identifiable from the sentence.
          The correct referent should be different in the two sentences.

    \item Both sentences in the pair must be read in a natural way, i.e. similar to what appears in common sources of text such as news articles and Wikipedia.
\end{enumerate}

Previous work modelled an ambiguous word as an observable, but they had two or more of them. In our case, we have ambiguous pronouns, but only one. In order not to end up with a scenario with only one observable, we define an observable to be a pair: (\textbf{pronoun}, \textit{special word}) or (\textbf{pronoun}, \textit{alternate word}). The possible outcomes of each of these observables are the candidate referents of the pronoun.


\begin{definition}[Winograd Schema scenario]
    Given a Winograd Schema with two noun phrases A and B; an ambiguous pronoun \textbf{p} which refers to either A or B; a special word (\textit{s}) and an alternate word (\textit{a}),
    the corresponding measurement scenario is defined by the data:
    \begin{itemize}
        \item observables $X = \{ (\textbf{p}, \textit{s}), (\textbf{p}, \textit{a}) \}$;
        \item contexts $\mathcal{M} = \bigl \{ \{(\textbf{p}, \textit{s})\}, \{(\textbf{p}, \textit{a})\} \bigr \}$;
        \item outcomes $O = \{\text{A}, \text{B}\}$.
    \end{itemize}
    Such a measurement scenario is called a \emph{Winograd Schema scenario}, or a WS scenario in short.
\end{definition}

With the \textit{trophy-suitcase} example, the measurement scenario would be given by the data:
\begin{itemize}
    \item observables $X = \{ (\textbf{it},\ \textit{large}),\ (\textbf{it},\ \textit{small}) \}$;
    \item contexts $\mathcal{M} = \bigl \{ \{(\textbf{it},\ \textit{large})\},\ \{(\textbf{it},\ \textit{small})\} \bigr \}$;
    \item outcomes $O = \{\text{trophy},\ \text{suitcase}\}$.
\end{itemize}
One can see that there is no overlap between the contexts and thus the empirical model corresponding to the schema is deterministic. It is shown in~\cite{Dzhafarov2019} that deterministic systems are not contextual and thus WS scenarios are not either.

\subsection{Extending Winograd to Generalised Winograd}

\begin{figure}[t]
    \centering
    \fbox{
        \includegraphics[width=0.95\textwidth, trim=0.5cm 21cm 0.5cm 0.5cm, clip]{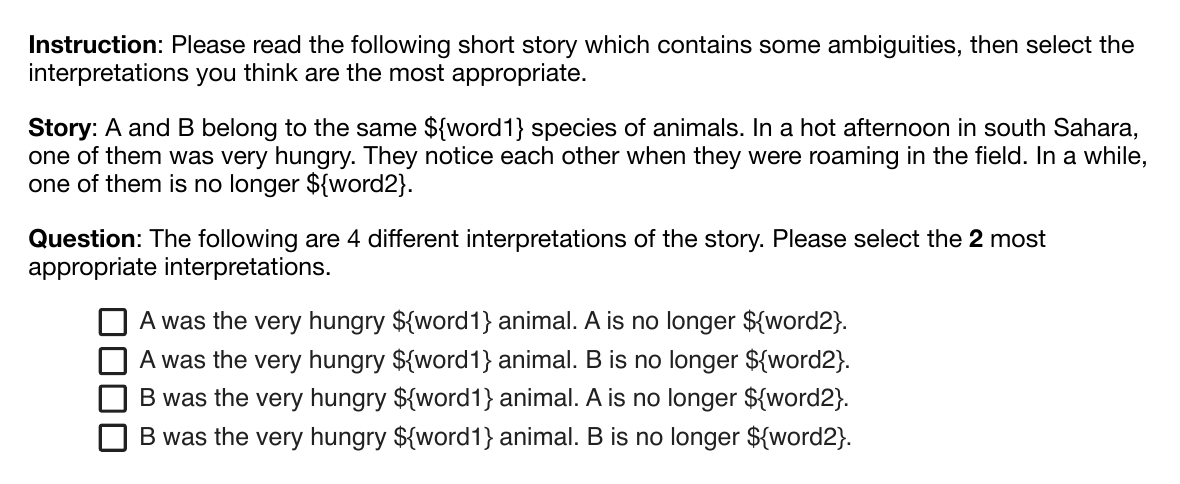}
    }
    \caption{A screenshot of the template of the questionnaire used to collect human judgments on the generalised Winograd Schema.}
    \label{fig:questionnaire}
\end{figure}
We generalised the Winograd Schema scenario such that contextuality can be exhibited in a recent work~\cite{Lo2023}.
The original WS is analogous to an imaginary physical experiment with only one experimenter, who decides whether to measure the pronoun with the special word, or with the alternate word, by choosing between the two observables: $(\textbf{p}, \textit{s})$ and $(\textbf{p}, \textit{a})$. A natural way of generalising it would be to introduce one more experimenter, resulting in the Bell-CHSH scenario.
This means that we need to add one more pronoun, one more special word and its alternate word. We use the subscript $1$ to denote components relating to the first pronoun and the subscript $2$ to those relating to the second pronoun. We obtain a new set of requirements for the resulting schema, which we call \emph{generalised Winograd Schema}:
\begin{enumerate}
    \item A generalised Winograd schema consists of four sentences. The first sentence contains two special words $\textit{s}_1$ and $\textit{s}_2$. Similar to the original Winograd Schema, $\textit{s}_1$ can be replaced by an alternate word $\textit{a}_1$, and $\textit{s}_2$ can be replaced by $\textit{a}_2$. Replacing special words with alternate words creates the rest of the four sentences.
    \item There are two pairs of noun phrases. The first pair can be identical to the second pair.
    \item There are two pronouns in the sentences. The first one refers to one of the noun phrases in the first pair of noun phrases. The second pronoun refers to either one of the noun phrases in the second pair of noun phrases.
    \item All four sentences should be natural to read.
\end{enumerate}

\begin{definition}[Generalised Winograd Schema scenario]

    Given a Generalised Winograd Schema with two noun phrases A and B; two ambiguous pronouns \textbf{p}$_1$ and \textbf{p}$_2$ can each refers to either A or B; two special words (\textit{s}$_1$) and (\textit{s}$_2$); two alternate words (\textit{a}$_1$) and (\textit{a}$_2$), the corresponding measurement scenario is defined by the data:
    \begin{itemize}
        \item observables $X = \{(\textbf{p}_1, \textit{s}_{1}), (\textbf{p}_1, \textit{a}_{1}), (\textbf{p}_2, \textit{s}_{2}), (\textbf{p}_2, \textit{a}_{2})\}$
        \item contexts $\mathcal{M} = \bigl\{ \{ (\textbf{p}_1, \textit{s}_{1}), (\textbf{p}_2, \textit{s}_{2}) \}, \{ (\textbf{p}_1, \textit{s}_{1}), (\textbf{p}_2, \textit{a}_{2}) \}, \{ (\textbf{p}_1, \textit{a}_{1}), (\textbf{p}_2, \textit{s}_{2}) \}, \{ (\textbf{p}_1, \textit{a}_{1}), (\textbf{p}_2, \textit{a}_{2}) \}\bigr\};$
        \item outcomes $O = \{\text{A}, \text{B}\}$.
    \end{itemize}
    Such a measurement scenario is called a \emph{Generalised Winograd Schema scenario}, or a generalised WS scenario in short.
\end{definition}
The generalised WS scenario is isomorphic, i.e.\ identical upon relabelling, to the Bell-CHSH scenario and should thus be able to host contextuality.

\section{Results}

\subsection{Basic Anaphoric Ambiguities revisited} 
\label{subsec:CbDbasicAnaphora}
As discussed in Section \ref{subsec:prprism}, the criterion of~\cite{Emeriau2022} is sufficient but not necessary.
To address this issue, we reanalysed the basic anaphoric ambiguities dataset using the CbD framework~\cite{Dzhafarov2013,Dzhafarov2016a,Dzhafarov2015a}, which provides a tight criterion for contextuality.
Indeed our analysis revealed that 9,159 examples (82.87\%) were contextual in the CbD framework.
Distributions of the signalling fractions and the Direct Influence of the CbD framework are shown in Figure~\ref{fig:nosigfracadj}. 

\begin{figure}[h]
    \centering

    \histogram{(0.0000, 7) (0.0417, 63) (0.0833, 100) (0.1250, 180) }{(0.1667, 223) (0.2083, 295) (0.2500, 376) (0.2917, 391) (0.3333, 397) (0.3750, 707) (0.4167, 489) (0.4583, 396) (0.5000, 716) (0.5417, 748) (0.5833, 615) (0.6250, 502) (0.6667, 580) (0.7083, 549) (0.7500, 625) (0.7917, 545) (0.8333, 798) (0.8750, 723) (0.9167, 723) (0.9583, 304) }
    \cbdhistogram{(0.0000, 106) (0.2500, 549) (0.5000, 1020) (0.7500, 1410) (1.0000, 1713) (1.2500, 1527) (1.5000, 1517) (1.7500, 1479) }{(2.0000, 512) (2.2500, 370) (2.5000, 339) (2.7500, 229) (3.0000, 144) (3.2500, 74) (3.5000, 37) (3.7500, 18) (4.0000, 7) (4.2500, 1) (4.5000, 0) (4.7500, 0) (5.0000, 0) (5.2500, 0) (5.5000, 0) (5.7500, 0) }
    \caption{Distributions of 11,052 examples of basic anaphoric ambiguities. The top histogram shows the distribution of the signalling fractions. 
    We observed that 350 examples (3.17\%) have a signalling fraction less than $1/6$, which is the threshold for conclusive contextuality according to the criterion of~\cite{Emeriau2022}.
    The bottom histogram shows the distribution of the Direct Influence of the CbD framework. We observed that 9159 examples (82.87\%) have a Direct Influence of less than $2$, which is the threshold for contextuality in the CbD framework.}
     
    \label{fig:nosigfracadj}
\end{figure}
\begin{figure}[t!]
    \centering
    \includegraphics[width=0.6\textwidth]{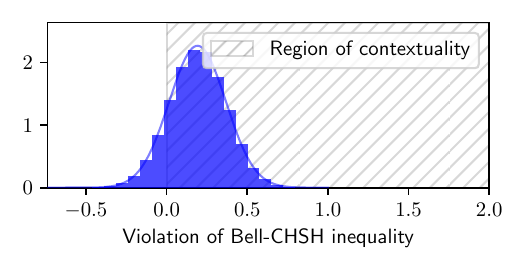}
    \caption{A histogram of violation of Bell-CHSH inequality
        for 100,000 bootstrap samples. }
    \label{figure:histo}
\end{figure}

\subsection{Human Judgments on the Generalised Winograd Schema}
We constructed an example of the generalised Winograd Schema and collected human judgments on this example on the crowd-sourcing platform Amazon Mechanical Turk in the form of a questionnaire. The example is:
\begin{quote}
    \it{ A and B belong to the same [\textit{cannibalistic} / \textit{herbivorous}]$_1$ species of animal. On a hot afternoon south of Sahara, \textbf{one of them}$_1$ was very hungry. They noticed each other when they were roaming in the field. In a while, \textbf{one of them}$_2$ is no longer [\textit{hungry} / \textit{alive}]$_2$.}
\end{quote}
There were four versions of the questionnaire, each corresponding to one of the four contexts in the generalised WS scenario.
The respondents were asked to read the example and answer a question about the correct referents, A or B, of the two referring phrases \textbf{one of them}$_1$ and \textbf{one of them}$_2$.
A screenshot of the questionnaire is shown in Figure~\ref{fig:questionnaire}. The placement holders \textsf{\$\{word1\}} and \textsf{\$\{word2\}} are instantiated with the two special words or the alternate words of the generalised Winograd Schema. In this example, \textsf{\$\{word1\}} can be either \textit{cannibalistic} or \textit{herbivorous} and \textsf{\$\{word2\}} can be either \textit{hungry} or \textit{alive}. Four versions of the questionnaire were created, each corresponding to one of the four contexts in the generalised WS scenario.

Since each referring phrase can be interpreted in two ways, there are 4 possible combinations of interpretations, (A, A), (A, B), (B, A), (B, B), of the two referring phrases.
The symmetry between A and B in the example ensures that the combinations (A, A) and (B, B) are equally plausible and (A, B) and (B, A) are also equally plausible.
Therefore we asked the respondents to pick two out of the four combinations.
This design choice also allows the detection of invalid answers, i.e.\ answers that are not symmetric with respect to A and B.

A total of 410 responses were collected on 20 Oct and 23 Nov 2022 from Amazon Mechanical Turk; 110 were for the context (\textit{cannibalistic}, \textit{hungry}) and 100 each for the rest of the three contexts. From these, 348 were valid. The respondents were each rewarded USD~1.00, regardless of the validity of their responses. The valid data was used to build an estimated probability distribution for each of the four contexts. The resulting empirical model is shown in Table~\ref{table:sahara}. The model violates the Bell-CHSH inequality by 0.192 with a standard deviation of 0.176 and is thus contextual. We conducted bootstrap resampling to establish the statistical significance of this result. The distribution of the violation of the resampled models is shown in Figure~\ref{figure:histo}. We see that of 87\% of the models have a positive violation with a standard deviation of 0.176. Our experimental model is thus contextual with a significance level of 87\%.

\begin{table}
    \centering

    \begin{tabular}{ll|ccccc}
        a)               &                  & (A, A) & (A, B) & (B, A) & (B, B) \\ \hline
        (\textit{canni}, & \textit{hungry}) & 0.402  & 0.097  & 0.097  & 0.402  \\
        (\textit{canni}, & \textit{alive})  & 0.044  & 0.455  & 0.455  & 0.044  \\
        (\textit{herbi}, & \textit{hungry}) & 0.345  & 0.154  & 0.154  & 0.345  \\
        (\textit{herbi}, & \textit{alive})  & 0.344  & 0.155  & 0.155  & 0.344  \\
    \end{tabular}
    \begin{tabular}{l|ccccc}
        b)    & (A, A) & (A, B) & (B, A) & (B, B) \\ \hline
        \dots & $1/2$  & $0$    & $0$    & $1/2$  \\
        \dots & $0$    & $1/2$  & $1/2$  & $0$    \\
        \dots & $1/2$  & $0$    & $0$    & $1/2$  \\
        \dots & $1/2$  & $0$    & $0$    & $1/2$  \\
    \end{tabular}
    \caption{
        (a) Empirical model constructed with human judgments from Amazon Mechanical Turk. The violation of Bell's inequality of the model is 0.192~$\pm$~0.176. For brevity, the special word \textit{cannibalistic} is shortened to \textit{canni} and the alternate word \textit{herbivorous} is shortened to \textit{herbi}. (b)
        Empirical model of the PR box.
    }
    \label{table:sahara}
\end{table}

\section{Discussion and Future Work}

We assumed that the alphabetic symbols $A,B$ used to model the generalised winograd scenarios are linguistic variables and thus interchangeable.
This had the advantage that the model became non-signalling and thus the contextual fraction remains a valid measure of contextuality.
Symmetrising psychological experiments has its criticisms, see~\cite{Cervantes2018}. We are, however, unaware of the existence of similar criticisms to a linguistic setting.
The symmetry in the outcomes allows the violation to saturate the bound defined by {\cf}~\cite{Abramsky2017} and the following equality is attained
\begin{align}
    \max\left\{0, \frac{1}{2}\ \textsf{violation of Bell-CHSH inequality} \right\} = \cf.
\end{align}
The CbD contextuality measures \cnt{1} and \cnt{2} coincide with the above degree of violation~\cite{Kujala2019}.
Thus, our model is considered contextual in both the sheaf-theoretic framework and the CbD framework.

The approach presented in this paper consists of deliberately constructing sentences that exhibit contextuality. One may criticise this as producing unnatural text. In future work, we will find naturally occurring language data that exhibits contextuality with the help of state-of-the-art generative large language models such as GPT-4~\cite{GPT4}.

\section*{Acknowledgements}
We are grateful to Daphne Wang for insightful discussions and the anonymous reviewers for their constructive comments.
KL is supported by the Engineering and Physical Sciences Research Council [grant number EP/S021582/1].
MS is supported by the Royal Academy of Engineering research chair RCSRF2122-14-152 on Engineered Mathematics for Modelling Typed Structures.

\bibliographystyle{eptcs}
\bibliography{references.bib}

\end{document}